# An Intelligent Remote Sensing Image Quality Inspection System

Yijiong Yu [1, 2], Tao Wang [2, *], Kang Ran [2], Chang Li [2] and Hao Wu [2]

[1] Department of Electronic Engineering, Tsinghua University; yuyj22@mails.tsinghua.edu.cn
[2] Guizhou Surveying and Mapping Product Quality Supervision and Inspection Station, Guiyang 550000, China
* Correspondence: wangt.07b@igsnrr.ac.cn; Tel.: +86-136-0855-3930

**Abstract:** Due to the inevitable presence of quality problems, quality inspection of remote sensing images is indeed an indispensable step between the acquisition and the application of them. However, traditional manual inspection suffers from low efficiency. Hence, we propose a novel deep learning-based two-step intelligent system consisting of multiple advanced computer vision models, which first performs image classification by SwinV2 and then accordingly adopts the most appropriate method, such as semantic segmentation by Segformer, to localize the quality problems. Results demonstrate that the proposed method exhibits excellent performance and efficiency, surpassing traditional methods. Furthermore, we conduct an initial exploration of applying multimodal models to remote sensing image quality inspection.

**1. Introduction**

    In recent years, rapid advancements in remote sensing technology have enabled the acquisition of large amounts of high-resolution remote sensing imagery, which has been widely applied in geological exploration, environmental monitoring, agriculture, and other fields. However, it is inevitable that remote sensing image quality problems will arise, which refer to the various factors that affect the accuracy, clarity, and reliability of remote sensing images. These issues are produced by multiple sources, including sensor limitations, atmospheric conditions, geometric distortions, radiometric errors, and transmission or compression artifacts. The presence of these quality problems in remote sensing images can compromise the interpretation and analysis of the acquired data, leading to inaccurate results and unreliable information.

    Therefore, remote sensing image quality inspection is necessary, whose ultimate purpose is to ensure the reliability and accuracy of remote sensing data and thus support informed decision-making and facilitate effective analysis and interpretation of the data, which is the same as remote sensing image quality assessment's. However, it has to be mentioned that quality inspection is not equal to quality assessment. The differences between them are shown in Table 1. The former can be seen as a more advanced and precise form of the latter, but also more demanding, which leads to very few researches on the automation of the remote sensing image quality inspection. Overreliance on experts' manual labor also limits the large-scale application of quality inspections to remote sensing data.

Table 1. Comparisons between quality assessment and quality inspection of remote sensing images.

|  | **Quality Assessment** | **Quality Inspection** |
| --- | --- | --- |
| **Goal** | According to the image characteristics, the quality of remote sensing images is scored to help screen out high-quality images. | Accurately mark the distortion area of the image, so as to provide a clear basis for whether the image is qualified and how the image producer should re-correct the image. |
| **Distortion range** | The distortion usually affects the whole image or the whole batch of images. | The distortion exists in some local regions of the image. |
| **Input** | the image to be evaluated, with or without the reference image | only the image to be evaluated |

| | | | |
|---|---|---|---|
| **Output** | Scores on one or more quality indicators | | The annotation box for each distortion area |
| **Existing common methods** | Subjective assessment: NIIRS evaluation method with GIQE equation [1]; Objective assessment: methods based on statistical features and machine learning models | | Most methods rely on visual examination by experts. |
| **Difficulty** | relatively simple and quick | | quite hard and time-consuming |

In our work, all the remote sensing images used default to RGB format with only visible spectrums. A distortion part of the image is usually called an "error" or a "quality problem", while the normal parts is called "qualified area" or "qualified regions". Figure 1 shows 10 common distortion types in remote sensing image quality inspection, which are visible to the naked eye, but difficult to describe mathematically in most cases.

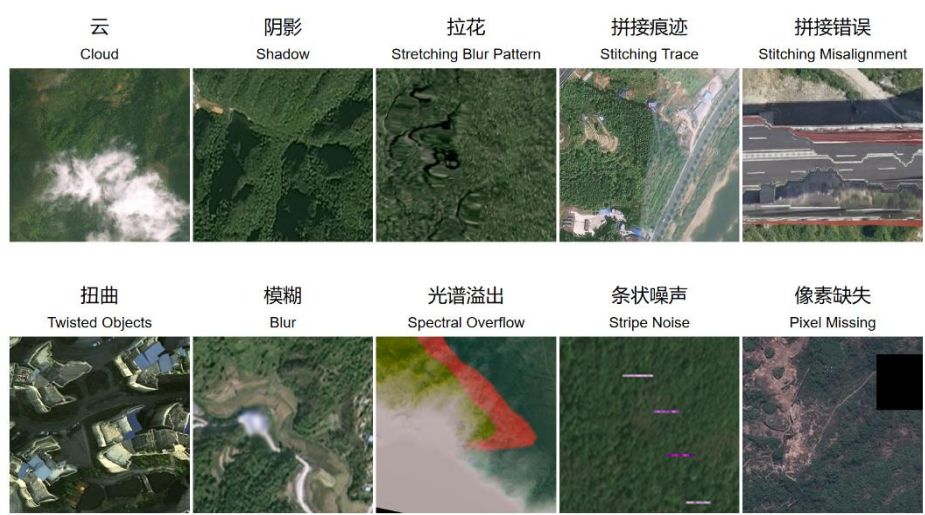

**Figure 1.** Example images of 10 distortion types and corresponding Chinese and English names. (We have provided both Chinese and English names because the translation of these distortion types is being introduced for the first time in this paper, and they do not have an official standard English translation.)

The task of remote sensing image quality inspection involves first identifying and marking regions within the image that do not meet the national quality standards and then calculating the percentage of the total image area covered by these regions. Finally, a decision is made on whether the image can pass the quality inspection. The low-quality images will be regarded substandard, and the corresponding image manufacturer will be required to repair or reshoot them. Figure 2 shows how we do remote sensing image quality detection with some examples.

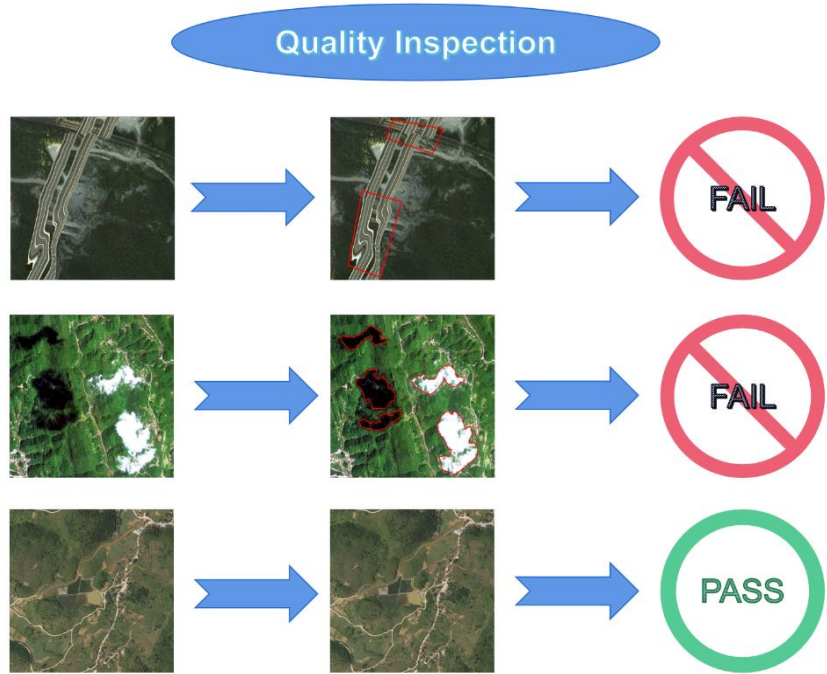

**Figure 2.** The work process of remote sensing image quality detection.

For this work, over the past decade, some automated detection methods have been proposed and achieved good results, such as object-based cloud detection methods [2], but such methods can only target one specific kind of objects. Therefore, it is of great significance to develop an efficient and universal method.

Because the work is to outline specific areas in the image, we treat this work as an image segmentation task in our study. Image segmentation is a time-honored subject, in which the most commonly used method is threshold-based algorithms, as well as clustering methods, edge-based methods, region-like methods, graph-based methods [3] and so on. However, these traditional image segmentation methods rely on manually set rules and parameters, making it impossible to deal with the complex distortions in remote sensing images, which are difficult to describe by mathematical rules. Hence, deep learning techniques, which can learn from data without the need for manually designed rules, have naturally become our preferred approach.

Deep learning techniques have made remarkable achievements in the field of computer vision, demonstrating powerful capabilities in image classification, object detection, semantic segmentation, and other tasks. In the field of image classification, the representative CNN-based [4] models are AlexNet [5], ResNet [6], DenseNet [7], MobileNet [8], etc., and in the image segmentation field, they are FCN [9], Unet [10], etc. Transformer-based [11] models, such as Swin [12] for image classification and Segformer [13] for semantic segmentation, are more advanced, and are gradually replacing the traditional CNN models.

Mainly based on deep learning models, we propose an intelligent two-step detection system, which is shown in Figure 3. Each image input to this system is first divided into smaller blocks, and an image classification model, SwinV2 [17] is used to determine the presence of the aforementioned 10 distortion types in each block. Subsequently, based on the distinctive characteristics of them, we categorize 10 distortion types into 3 major classes and design a specific detection method for each, which will be introduced in detail in section 2. Finally, the detection results of each block and each distortion type are fused into one. The experimental results confirm the superiority of our two-step approach compared to previous methods.

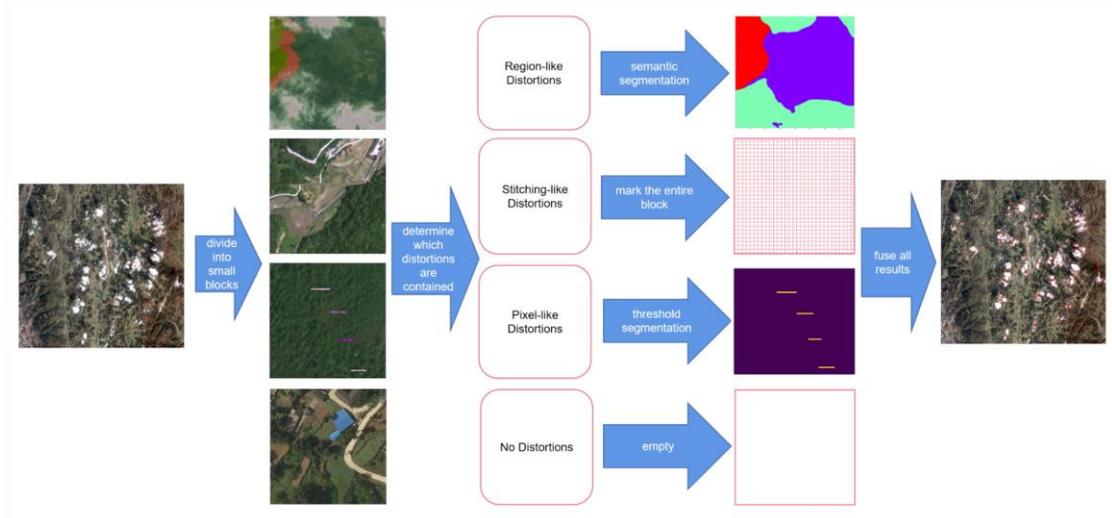

**Figure 3.** The process of the two-step system is presented with real remote sensing images. (The pictures of the blocks are not actually from the same remote sensing image.)

Although our two-step system has achieved good results, it can only be used to detect these fixed types of distortions. In contrast, universal large models like GPT-4 [14] will provide users with more convenient customizable functions. Based on this intention, we explores the role of image-text multimodal models such as BLIP [15] and VisualGLM [16] in remote sensing image quality inspection, hoping these multimodal models can provide more comprehensive, detailed, and customizable descriptions about the distortion in the remote sensing image, such as areas, locations, colors and shapes, to some extent replacing image classification models and segmentation models.

In summary, our main contributions are as follows:

1. We annotate a dataset for remote sensing image quality inspection, and to our knowledge, this is the first publicly available dataset in this field. Appendix A describes how we collect the data and make the dataset.

2. We perform a specific analysis for each distortion type, categorize them into three major classes, and design the most suitable detection method for each major class.

3. We decompose the remote sensing image quality inspection task into multiple subtasks. In each subtask, we select the most suitable model and method through comparative experiments, and thereby integrate them to construct the complete two-step quality inspection system.

4. We conduct preliminary experiments on applying the multimodal question-answer model to image quality inspection, which demonstrates its potential for future applications.

## 2. Methods

The flowchart of the two-step detection system for remote sensing image quality inspection is shown in Figure 4 which has a branch structure. Why we adopt such a structure and how to implement it will be detailed in this section.

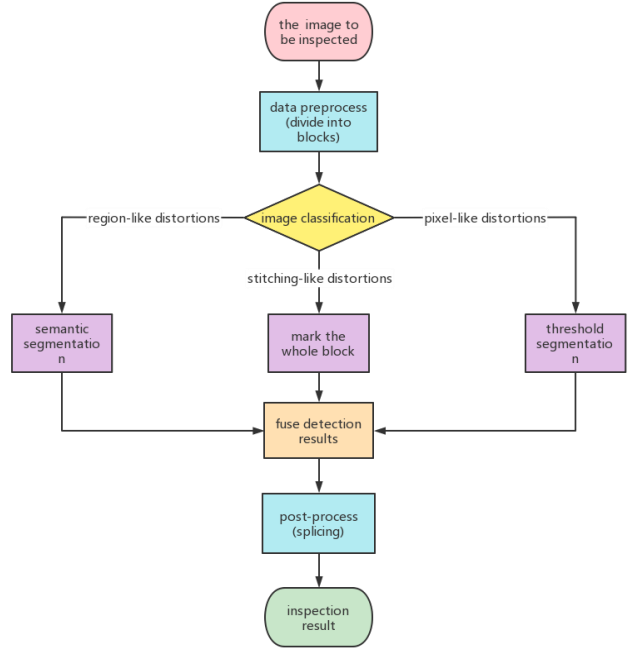

**Figure 4.** The flowchart of the two-step detection system.

*2.1. Why Two-step is Necessary?*

Our approach incorporates an additional step of image classification before identifying the specific locations of distortions, unlike previous distortion detection methods that typically involved a single step. The use of our two-step detection process offers the following purposes and advantages:

2.1.1. To Apply Specific Methods to Specific Problems

In our workflow, the second step employs a branch structure, enabling the selection of specific detection methods for different types of distortions. The image classification step in the first stage provides the foundation and prerequisite for selectively applying these methods.

Also, this structure maintains scalability when faced with an increasing number of potential distortion types in the future. For new types, it only requires adding a new branch and retraining the image classification model, without the need to modify the existing branches.

2.1.2. To Reduce Computational Cost

Due to the requirement of different methods for different types of distortions, if the entire image is to be examined, at least three methods need to be applied to the whole image, and each pixel needs to be involved in 3 computations, resulting in a significant computational burden. On the contrary, because based on experience, the area of distortion typically accounts for no more than 30% of the image being examined, by using a two-step approach, where classification is performed first, most blocks will be identified as "no error" and do not require participation in the second step computation. This greatly reduces the computational workload. Therefore, on average, each pixel needs to be involved in a maximum of $1 + 3 * 0.3 = 1.9$ computations, resulting in a lower computational burden.

2.1.3. To Improve Precision

Semantic segmentation is a task of assigning every pixel a class label of its corresponding object. However, simply applying one semantic segmentation model would lead to a terrible performance caused by the characteristic of our task—class-imbalance. As mentioned earlier, the qualified region typically accounts for more than 70% of the image area. If we train a semantic segmentation model without selecting specific regions from remote sensing images, the number of negative samples (blocks containing only qualified regions) will far exceed the number of positive samples (blocks containing distortion regions), resulting in a low recall score for the trained model. A weighted loss function may help but cannot fully address this issue. On the contrary, if we train using only positive samples and solely rely on semantic segmentation models, we will encounter a significant amount of scattered and small false detection noise in regions where distortions do not actually exist. This leads to unreliable detection results and low precision. We seem to be caught in a dilemma.

Fortunately, image classification models are relatively less affected by such imbalanced situations. In our experiments, we find that even when the negative samples account for approximately 50% of the training data, the accuracy of the image classification model remains good. However, semantic segmentation models tend to classify all pixels as "no error," rendering the model unusable. Therefore, in order to ensure a high recall rate, we do not use negative samples when training the semantic segmentation model. However, for the image classification model, we do incorporate negative samples to enable it to differentiate between positive and negative instances.

Then, by conducting image classification as a preliminary screening step, the distortion types that do not appear in the classification results will not be present in the final detection results, which means image classification effectively suppresses false detection noise caused by semantic segmentation or threshold segmentation, significantly improving precision.

2.1.4. To Handle Distortions with Indefinable Boundaries

Based on our current research, both human experts and computers find it challenging to accurately determine the exact seams of stitching errors, because these distortions usually manifest themselves as object discontinuity or color difference, but the shape, size, color, and potential occurrence locations of the boundaries are irregular. Currently, deep learning models can determine whether the given block contains stitching errors, but we have not yet found a model that can precisely locate such stitching seams. That is why the image classification step is necessary—it plays a role as a rough localization of these types of distortions.

*2.2. Data Preprocessing*

In the field of remote sensing, original images are often too large to be processed directly by deep learning models. Therefore, a preprocessing step is necessary to divide these images into smaller blocks. Instead of a simple division approach, we propose a specialized division strategy that incorporates two key features:

1. Overlapping Blocks: To address the issue of incomplete information and potential false negatives caused by dividing a distorted region into two blocks, our method employs blocks with overlapping areas. Specifically, we set the width of the overlapping pixels between blocks to approximately 1/5 of the image's side length. This ensures that important details are not lost during the division process.

2. Variable Block Sizes: Defective regions in remote sensing images, such as "cloud", can vary significantly in size, ranging from tens of meters to tens of kilometers. Detecting both large and tiny distortions effectively can be challenging when dividing the original image into fixed-size blocks. To overcome this challenge, we adopt a multi-scale approach. For each image, we divide it into multiple sizes of blocks, typically using two sizes: $512*512$ and $1024*1024$ pixels. This allows us to perform accurate inspections at different scales, effectively capturing distortions of various sizes.

*2.3. The First Step: Image Classification*

To enable the model to determine that a block has no distortion, the dataset used for image classification tasks has approximately 30% of the samples representing images with no distortion.

For image classification, we utilize advanced pretrained models such as Swin Transformer V2 model [17] to determine which types of distortions are contained in each block. This classification task involves multi-label binary classification. The last classification layer's output dimension of the pretrained model is set to 10, while the other parts of the model remain unchanged. Given an input image, the model generates a 10-dimensional vector representing the logits for the 10 distortion types. (Logits are the network's output values before any activation function is applied.) We employ the binary cross-entropy loss function, as shown in equation (1), where $\hat{y}$ represents the 10-dimensional network output vector and $y$ denotes the true label, which is also a 10-dimensional vector with elements of 1 or 0. $C$ denotes the number of categories, in our study it is 10. $y_c$ represents the c-th dimension of $y$.

$$Loss = -\frac{1}{C}\sum_{c=1}^{C}\left(\hat{y}_c \log \sigma(y_c) + (1-\hat{y}_c)\log(1-\sigma(y_c))\right) \tag{1}$$

When converting logits into classification results, a common threshold of 0.5 is applied. If the sigmoid activation value of a certain distortion type's logit exceeds 0.5, it is considered as present; otherwise, it is considered absent. If all logits are less than 0.5, the image is deemed fully qualified (no distortion).

*2.4. Three Major Class of Distortions*

The 10 distortion types are classified into the following three major classes:

**Region-like distortions:** Clouds, Shadows, Stretching Blur Patterns, Twisted Objects, Blurs, and Spectral Overflow fall under this category. The distortion regions exhibit region-like shapes with a certain area. Quality inspection of remote sensing images needs to outline these regions and calculate their areas. Thus, semantic segmentation methods are competent.

**Stitching-like distortions:** Stitching Traces and Stitching Misalignments belong to this category. The distortion regions exhibit stitching-like patterns, often spanning several kilometers. To identify the seam, a comparison between two adjacent regions with a larger visual range is required. It has been mentioned earlier that precise detection of the exact seam location is hard, so we choose to just mark the whole block where the seam line is located by image classification methods. Thus, the detection effect of such distortions will not be evaluated separately, because it will be reflected in the evaluation of the image classification procedure.

**Pixel-like distortions:** Pixel Missing and Stripe Noise belong to this category. The distortion regions only occupy a few pixels, with extremely tiny areas. Semantic segmentation is not suitable for them due to their tiny size. However, they possess distinct color characteristics, making them detectable using threshold segmentation methods.

*2.5. Handling Region-like Distortions: Semantic Segmentation*

For region-like distortions, there is subjectivity in the segmentation process between "instances", in other words, it is challenging to precisely define the size of an "instance". Consequently, using instance segmentation or panorama segmentation can confuse the model during training. Instead, it is more natural to employ semantic segmentation models that do not require distinguishing instances.

We experiment with 4 semantic segmentation models: MobilenetV2 [18], Segformer [13], CLIPSeg [19], and GroupVit [20]. The first two are classic semantic segmentation models, while the latter two are text-driven segmentation models. Although models such as Mask2Former [21] and OneFormer [22] are more advanced in terms of semantic segmentation performance theoretically, we experimentally find that they are better suited for instance segmentation or panorama segmentation, but have larger model sizes and slower training speeds, which means they are not very cost-effective. Therefore, they are not chosen in this study.

Since adding negative samples (fully qualified images) would result in an excessive number of background pixels in the training dataset, which impairs the performance of the semantic segmentation model, we keep only the positive samples in the training dataset. The loss function used to train MobilenetV2 [18] and Segformer [13] is the average of multi-class cross-entropy for all pixels, as shown in Equation 2, where $K$ is the total number of pixels in one image, $C$ is the number of categories, $y_c^{(k)}$ is the predicted logit of $k$-th pixel of class $c$ while $\hat{y}_c^{(k)}$ is the binary label, and $\sigma$ represents the $softmax$ function. For CLIPSeg [19] training, the loss function is the average binary cross-entropy for all pixels in each binary mask, which can be seen as a special case where $C = 2$ in Equation (2). GroupVit [20], on the other hand, does not require masks as labels but uses the textual description of the image as the label. The loss function for GroupVit [20] is contrastive loss, whose further details can be found in the original paper [20].

$$Loss = -\frac{1}{K}\frac{1}{C}\sum_{k=1}^{K}\sum_{c=1}^{C} \hat{y}_c^{(k)} \log \sigma\left(y_c^{(k)}\right) \tag{2}$$

The segmentation results produced by the semantic segmentation model will be adjusted based on the image classification results from the previous step. This will be discussed in detail in section 2.6.

*2.6. Integration Methods of Image Classification and Semantic Segmentation*

2.6.1. Constraint at the Output Side

For the models having traditional semantic segmentation headers such as Segformer [13], the classification model can constrain their outputs in three possible ways:

**Absolute constraint:** The output logits of the image classification model are simply converted into binary classification results. If a distortion category is not present, the corresponding pixels associated with that distortion type are discarded in the semantic segmentation results, i.e., set to 0. This method is advantageous in terms of convenience and computational speed. However, it can lead to error accumulation. If the image classification model provides incorrect predictions, even if the semantic segmentation has correctly detected the distortion, it may still be discarded, resulting in a high probability of missed detections.

**Relative constraint:** The logits of the corresponding distortion type in the semantic segmentation output are adjusted based on the confidence scores of each distortion type in the image classification results. Specifically, the logits of distortion type "A" in the semantic segmentation output are added to the logits of distortion type "A" in the image classification results. This method ensures that the logits from both models are considered. However, changing the logits of certain categories may lead to unstable semantic segmentation results. For example, if the confidence score for "shadow" is high in image classification, it may result in nearly all pixels being classified as shadows in the semantic segmentation results.

**Threshold-based constraint:** This method involves setting a threshold, typically 0.5. If the logits for distortion type "A" in the image classification results exceed the threshold, no adjustments are made to the logits in the semantic segmentation corresponding to that distortion type. Otherwise, the logits of distortion type "A" in the semantic segmentation output are added to the logits of distortion type "A" in the image classification results after multiplied by a coefficient. This method offers the advantages of only lowering the confidence of categories with lower scores, while minimally affecting categories with high confidence scores. It mitigates the impact of image classification results on semantic segmentation and ensures that distortion types with low confidence in image classification can still be detected to some extent, reducing the missed detection rate.

These three constraint methods can be represented by one equation (3):

$$\hat{y}_s^{(c)} = \alpha y_c^{(c)} \epsilon \left(t - y_c^{(c)}\right) + y_s^{(c)} \tag{3}$$

$\alpha$ is a coefficient and $t$ is a threshold (usually 0.5), which can be adjusted by users. $\epsilon$ stands for Heaviside step function. $y_c^{(c)}$ represents the logit of class $c$ in the predicted results of the classification model. $y_s^{(c)}$ represents the logit of class $c$ of each pixel predicted by the segmentation model and $\hat{y}_s^{(c)}$ is the corresponding adjusted logit. When $\alpha$ approaches positive infinity and the threshold is set to 0, it corresponds to the absolute constraint. When the threshold approaches positive infinity and $\alpha = 1$, it corresponds to the relative constraint. Setting other values can be seen as threshold-based constraint.

2.6.2. Constraint at the Input Side

For models such as CLIPSeg [19] and GroupVit [20], which are text-driven zero-shot semantic segmentation models, both the image and a list of $n$ target object names are required as inputs. Therefore, the classification model provides constraints at the input side. The specific approach is as follows: First, the image classification results are transformed into n texts, where each text corresponds to the name of a distortion type classified as "yes." For example, if the "cloud" and "shadow" classes are classified as "yes" in the classification results, the input texts would be ["cloud", "shadow"]. These texts are then combined with the image and fed into the model. The model performs binary segmentation for each object mentioned in the input texts and does not perform segmentation for distortion types not present in the classification results, thereby avoiding false detections.

*2.7. Handling Pixel-like Distortions: Threshold-based Segmentation*

For blocks that exhibit pixel-like distortions, semantic segmentation is not suitable due to their tiny size. However, these blocks exhibit distinct pixel values compared to the background. Therefore, we employ traditional threshold-based image processing methods for discrimination.

Regions with Pixel Missing typically manifest as pixels that are either completely black or white, without any specific shape. To detect such regions, a threshold-based segmentation method is applied. Specifically, regions with pixel values of [0, 0, 0] or [255, 255, 255] are directly extracted as the detection results.

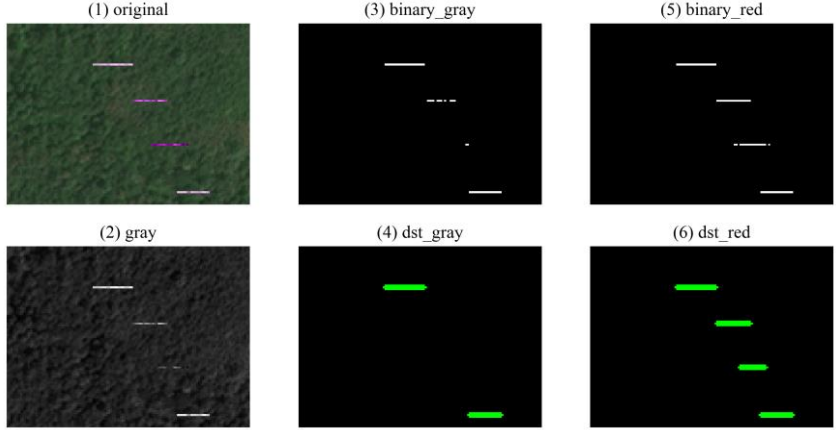

**Figure 5.** (1) Original image, (2) grayscale image, (3) grayscale image binarization, (4) result of extracting stripe noise from the binarized grayscale image, (5) the red channel binarization, (6) result of extracting stripe noise from the binarized red channel.

Stripe-like Noise exhibits distinct shape characteristics, typically manifesting as short, thin lines that are noticeably different in color from the background. These lines are predominantly oriented horizontally or vertically, occasionally diagonally, but seldom appear as curves or jagged lines. Exploiting their shape and color features enables effective extraction of such noise. Typically, grayscale conversion is performed on images prior to binarization. However, we find in the experiment that in the grayscale images, stripe-like noise and the background may exhibit considerable similarity, while in a specific color channel, the noise displays a distinct color contrast to the background. The results of binarization using the grayscale image and the red channel, along with the extracted stripe-like noise, are illustrated in Figure 5. It is evident from the figure that employing binarization on the red channel yields significantly superior detection results compared to binarizing the grayscale image.

Hence, we employ the following approach for stripe-like noise detection: RGB channels are first separated, and Otsu binarization [23] is applied to each channel to detect connected components. Subsequently, the shape of all connected components is analyzed, and any component that appears as a rectangular region with an aspect ratio greater than 5 or less than 0.2 is identified as stripe-like noise. Finally, the union of the detection results of the three channels will be taken.

*2.8. Data Post-processing*

When merging and stitching the detection results of different blocks, conflicts may arise. A "conflict" refers to the situation where there are two or more different distortion types detected at the same position in the original image by different blocks. In case of a conflict, the region-like distortion will be determined based on the confidence score output by the semantic segmentation model. The priority order between 3 major distortion types is as follows: "pixel-like" distortions first, followed by "region-like" distortions, and finally "stitching-like" distortions. Distortions of high priority will override those of low priority.

Furthermore, to reduce noise in the results and avoid scattered and small detection areas, a dilation operation is performed on the binary mask corresponding to each distortion type in the semantic segmentation results. Then, connected components with an area smaller than a threshold (usually set to 0.01%) of the image area are discarded.

The concatenated mask, using different pixel values to represent different distortion types, is a single-channel image with the same size as the original remote sensing image. OpenCV is used to perform contour detection on this mask, describing each distortion region as a polygon, and finally convert it to a shapefile.

*2.9. Exploration of Multimodal Models*

In addition to traditional fixed-label image classification, the recently proposed image-text multimodal models can also be used for image classification. Although their classification accuracy may not be as high as dedicated classification models for specific tasks, they provide more comprehensive information and have greater versatility. They hold the potential for "one model for all tasks" and may be capable of addressing all image-related problems in remote sensing. Therefore, the application of multimodal models in the field of remote sensing is promising.

The structure of multimodal models is similar to generative language models, but with the addition of an image encoder that embeds the image as a vector alongside the textual prompt. The output of the model is identical to that of a language model. The training methodology for multimodal models is largely similar to language models and will not be elaborated upon in this paper.

A comparative experiment is conducted between BLIP [15], which possesses the ability to convert images into text, and VisualGLM [16], which supports long conversations. We do not need to make any structural changes to the pretrained models. For BLIP [15], a Chinese pre-trained version is utilized, and the dataset mentioned in Appendix A.2., consisting of images and their corresponding textual descriptions, is used for fine-tuning all parameters. As for the VisualGLM [16], due to its large size, a lora[24] training approach is employed. Only the layers with even numbers in the ChatGLM [16] part of VisualGLM [16] underwent lora training, while no modifications are made to other parts such as the image encoder. Subsequently, the Q&A dataset mentioned in Appendix A.4. is used to train the model's ability to provide detailed descriptions of distortions in remote sensing images.

## 3. Experiments

In this section, we first evaluate every individual process in the system, then the model integration method, and finally the multimodal model. Due to the large difference between the detection methods and result forms of different distortion types, it is hard to measure our whole system with a single metric, thereby, we only provide respective evaluations of the detection effect for each distortion type.

*3.1. Training Settings*

All models are trained on a single NVIDIA GeForce RTX 3090 GPU using the AdamW optimizer. The learning rate scheduler is set to "cosine" with an initial learning rate of 1e-4. Appendix B contains the download links, parameter counts, the training input, label, test input, and output configurations for each pretrained model.

For all models except VisualGLM [16], random data augmentation is applied to the input images during training, including random horizontal and vertical flipping and random cropping on both images and their corresponding masks. Gaussian noise augmentation is not used to avoid affecting the model's detection.

For all image classification models, training is conducted for 10 epochs, while semantic segmentation models are trained for 5 epochs. In the case of Segformer [13], the background (qualified area) is treated as an additional class with label 0, and negative samples (images without distortions) are excluded from the training set to avoid affecting the segmentation recall. The same approach is applied to CLIPSeg [19], where the background is also considered as a class. Correspondence between IDs and labels during models training is shown in Table 2.

**Table 2.** Correspondence between IDs and labels of the models.

| ID | Labels of Classification Models | Labels of Semantic Segmentation Models |
|---|---|---|
| 0 | Cloud | Background |
| 1 | Shadow | Cloud |
| 2 | Stretching Blur Pattern | Shadow |
| 3 | Blur | Stretching Blur Pattern |
| 4 | Spectral Overflow | Blur |
| 5 | Twisted Objects | Spectral Overflow |
| 6 | Stitching Trace | Twisted Objects |
| 7 | Stitching Misalignment | \ |
| 8 | Stripe Noise | \ |
| 9 | Pixel Missing | \ |

For the multimodal model VisualGLM [16], lora training is employed with a lora rank of 16, focusing on the ChatGLM part and performing lora on layers [0,2,4,6,8,10,12,14,16,18,20,22,24,26] for 5 epochs. The Unet part of the stable diffusion model is also trained using lora with a "lora_rank" of 10 for 10 epochs.

*3.2. Evaluation of Image Classification Models*

ResNet [6] and SwinV2 [17] models are evaluated for image classification using accuracy, macro F1, and F1 scores per class. The text output from the image-to-text model BLIP [15] is mapped to labels in the same format as the classification model to be evaluated. For example, if BLIP [15] outputs "Satellite image with shadows and blur," it is mapped to the label [0,1,0,1,0,0,0,0,0,0].

Table 3 presents the experimental results for the classification task on the dataset. The accuracy represents the proportion of correctly classified samples for all 10 classes. The results demonstrate that the SwinV2 [17] model achieves better performance in remote sensing image quality classification, nearly achieving perfect classification for all classes, while ResNet [6] consistently performs weaker than other models. The BLIP [15] model also exhibits strong classification capabilities.

Table 3. Accuracy and macro f1 of the image classification models.

| Model | Accuracy | Macro F1 | F1 of Each Category |
|---|---|---|---|
| Resnet-50 | 0.9192 | 0.8491 | [0.9719, 0.9701, 0.8947, 0.8125, 0.9734, 0.9600, 0.9600, 0.6000, 0.5000, 0.8500] |
| SwinV2-base | **0.9719** | **0.9827** | [0.9824, 0.9934, 0.9743, 1.0000, 0.8461, 1.0000, 1.0000, 0.9937, 1.0000, 0.9800] |
| BLIP | 0.9670 | 0.9611 | [0.9860, 0.9832, 0.9295, 0.9600, 0.9302, 1.0000, 0.8666, 0.9941, 1.0000, 0.9600] |

*3.3. Evaluation of Semantic Segmentation Models*

Semantic segmentation is evaluated using the mean Intersection over Union (mIoU) metric in a test dataset without negative samples. We compare 4 segmentation models, Mobilenet-v2 [18], Segformer [13], CLIPSeg, and GroupVit [20], and the rule-based method, whose rule is detailed in Appendix C. The input for CLIPSeg [19] and GroupVit [20] during testing included both the test images and the names of existing distortion types in the images. Table 4 presents the mIoU scores and IoU scores of each class on the test dataset (these 7 categories are Background, Cloud, Shadow, Stretching Blur Pattern, Blur, Spectral Overflow and Twisted Objects in order).

Table 4. IoU of the semantic segmentation models.

| Model | mIoU | IoU of Each Category |
|---|---|---|
| Rule-based | 0.16006 | [0.86193, 0.58395, 0.15472, 0.00000, 0.00000, 0.00000, 0.00000] |
| Mobilenet-v2 | 0.47153 | [0.91771, 0.81417, 0.26575, 0.38780, 0.17978, 0.41014, 0.32538] |
| Segformer | **0.71266** | [0.95451, 0.86814, 0.65336, 0.48435, 0.90058, 0.76419, 0.36348] |
| CLIPSeg | 0.40703 | [0.93582, 0.84138, 0.35900, 0.00941, 0.00000, 0.70027, 0.00330] |
| GroupVit | 0.19406 | [0.38743, 0.45659, 0.06993, 0.13790, 0.19186, 0.00175, 0.11297] |

The results indicate that Segformer [13] outperforms other segmentation models in remote sensing image semantic segmentation tasks. In contrast, the performance of text-driven models is poor, even when additional information, such as the names of existing distortion types in the images, is provided. This might be attributed to the need for extensive image-text data training for such models, while this study employed limited data and relatively homogeneous image categories. The IoU of each category shows that Clouds, Shadows, Blurriness, and Spectral Overflow are well segmented by Segformer [13], while distortions with more complex features such as Stretching Blur Pattern and Twisted Objects are still relatively hard to handle.

*3.4. Evaluation of Model Integration Methods*

To demonstrate the effect of the integration methods of the two models, pixel precision (each pixel as a separate classification task) and mIoU are used to evaluate the final performance of semantic segmentation on region-like distortions under different settings of parameter $\alpha$ and $t$. The two models integrated in this experiment are SwinV2-base [17] and Segformer-b5 [13]. The experiment results are shown in Table 5, which confirm that integrating two models significantly outperform using the semantic segmentation model alone (one-step), especially in terms of precision. More specifically, when $\alpha$ approaches infinity, indicating absolute priority given to the classification model, the overall performance is optimal, while setting $\alpha$ to 0, equivalent to the one-step method, leads to the poorest performance.

It can be more intuitive displayed in Figure 6 that, for the image with large distortion area, two-step detection is almost consistent with one-step detection, proving that the two-step detection, although more constrained, does not cause missing detection and maintains high recall, and for images without distortion regions, the results of one-step detection contain a large amount of noises of false detection, while two-step detection does not, demonstrating the high precision of the two-step detection.

It is worth noting that the negative samples in our test dataset only account for about 30%, which is much lower than that in the practical scenario, where our two-step method should achieve more significant superiority to one-step, especially in precision.

**Table 5**. The influence on semantic segmentation of $\alpha$ and $t$ in equation (3).

| $\alpha$ | $t$ | mIoU | Pixel Precision |
|---|---|---|---|
| 0 | 0 | 0.6904 | 0.7880 |
| 100 | 0.5 | **0.7124** | **0.8154** |
| 1 | 0.5 | 0.7123 | 0.7985 |
| 0.3 | 0.9 | 0.7120 | 0.7981 |
| 0.1 | 0.5 | 0.6980 | 0.7831 |
| 0.3 | 0.01 | 0.7122 | 0.7984 |

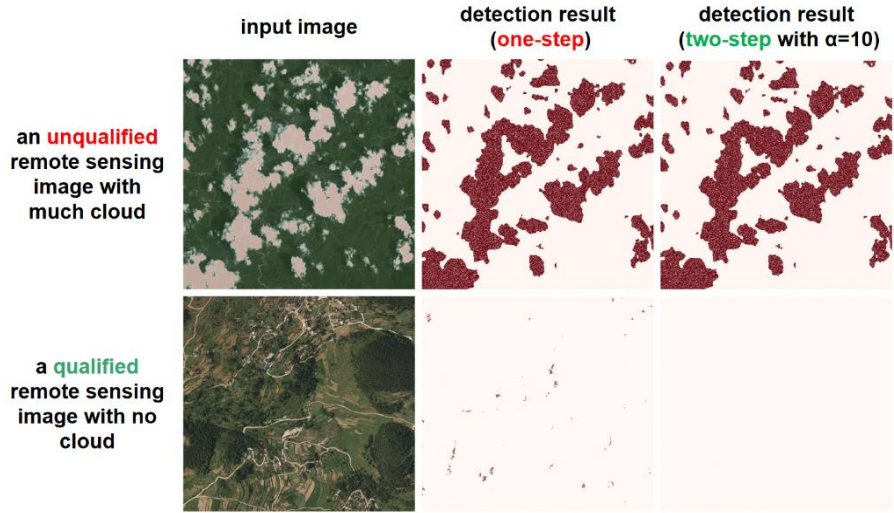

**Figure 6.** An unqualified image and a qualified image are tested at different values of $\alpha$ and $t = 0.5$. In the detection result figure, the dark parts represent the detected distortions and the light part represents the background.

*3.5. Detection Effect on Pixel-like Distortions*

For pixel-like distortions, which include only two distortion types so far, IoU of each is calculated likewise. The results are shown in Table 6, demonstrating good detection performance. In addition, it is found in the experiment that an appropriate saturation enhancement (in this experiment, the S channel value in HSV is increased by 100) can significantly improve the detection result of Stripe Noise. However, since these types of distortions are relatively scarce in the current test dataset, further verification is expected in the future to validate their effectiveness.

**Table 6.** IoU of the detection methods for pixel-like distortions.

| Detection Type | IoU |
|---|---|
| Stripe Noise Detection w/ saturation enhancement | 0.4996 |
| Stripe Noise Detection w/o saturation enhancement | 0.3860 |
| Pixel Missing Detection | 0.9763 |

*3.6. Evaluations of the Multimodal Model*

The VisualGLM [16] model trained on image-based Q&As is evaluated on a test dataset consisting of 2000 question-answer pairs in the same format. Accuracy and Rouge scores are used for evaluation. The evaluation results are presented in Table 7. Due to the relatively unpredictable outputs of the model, with answers that may be excessively long or not in the expected format, assessing classification accuracy directly becomes challenging, and thus the accuracy metric is divided into two parts: The first part (exact match) regards the model's output to be correct only if it matches the label exactly, while the second part (binary classification) regards so if the first word of the model's output ("yes" or "no") aligns with the label. The Rouge scores has three aspects, representing Rouge-1, Rouge-2, and Rouge-L evaluations.

The Q&A examples in Figure 7 and the evaluation results indicate that the fine-tuned VisualGLM [16] model can provide detailed descriptions of images and has the capability of image quality inspection. However, the accuracy is not competitive with traditional classification models, and there are still cases where the answers do not correspond to the image content or fail to address the question, suggesting room for further improvement.

**Table 7.** Evaluation results of VisualGLM.

| Model | Accuracy (Exact Match) | Accuracy (Binary Classification) | Rouge-1 F1 | Rouge-2 F1 | Rouge-l F1 |
| --- | --- | --- | --- | --- | --- |
| VisualGLM | 0.7144 | 0.8409 | 0.9000 | 0.8664 | 0.8847 |

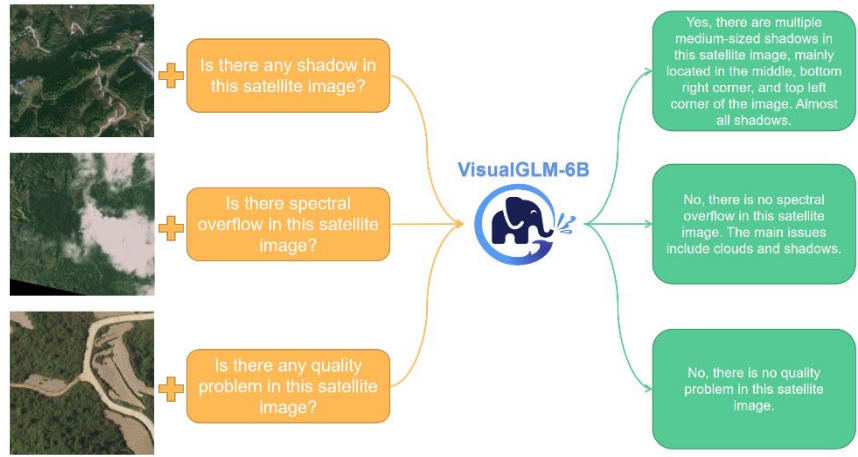

**Figure 7.** Some answers generated by VisualGLM given images and questions about image quality. (The actual model is trained for Chinese Q&A.)

**4. Discussion**

*4.1. The Practical Effect of Our System*

In the practical application of the system, we adopt SwinV2-base [17] as the classification model and Segformer-b5 [13] as the segmentation model. The block size is set to $512*512$ and $1024*1024$ and the overlap width is set to one-fifth of the block width, in most cases. $\alpha$ is set to 10 and $t$ is set to 0.5.

By applying this intelligent system to thousands of images in our practical work, we find that this method is more effective in detecting common distortions such as Cloud and Shadow, while due to the complex features and small amount of training data, Twisted Objects, Stretching Blur Pattern and stitching-like distortions are often omitted. This situation may be improved by further adding high-quality annotation data or designing a targeted model structure. The detection effect of pixel-like distortions is somewhat hard to tell, because their occurrence frequency is very low in practice, as well as in our dataset. Therefore, although our system has been evaluated in our test dataset, its power still needs to be further validated in practical work.

Fortunately, after checking the testing results from practical work, our experts commented that, although omissions often occur, this system has very few false detections, indicating a high level of precision.

By using this system to inspect images and then conducting manual screening to supplement any omissions by it, the overall efficiency remains significantly higher than relying solely on manual inspections.

It is worth noting that the remote sensing images used for training and testing in our study were all sourced from Guizhou Province, which is characterized by mountainous terrain, significant elevation changes, and frequent cloud and fog. This implies that quality inspection of remote sensing images here is more challenging compared to flat regions. However, our system still demonstrates satisfactory performance, indicating its robustness and potential for application in other countries and regions.

We also tested the time cost of our two-step method. The result shows this method takes about five minutes to detect an image at 10000 * 10000 resolution on an entry-level graphics card with only 4GB memory in a single-threaded manner, which is much more efficient and economical than humans. The time and memory required is roughly proportional to the total number of pixels in the image. Therefore, we would recommend against detecting an image larger than 1GB at a time.

*4.2. Limitations of Multimodal Models*

There are two types of multimodal models in this paper: text-driven segmentation models (CLIPSeg [19] and GroupVit [20]) and image-based language models (BLIP [15] and VisualGLM [16]). Ideally, the multimodal model should have better accuracy and higher flexibility, but in fact, both of they perform worse in our experiments in most cases.

In terms of image classification, the SwinV2 [17] model achieves the best score and BLIP [15] has a slightly lower accuracy, which proves that the multimodal model performs fairly well on a relatively simple classification task.

As for the semantic segmentation task, Segformer [13] achieves the best performance, while text-driven segmentation models like CLIPSeg [19] and GroupViT [20] lag far behind the former.

The multimodal large model VisualGLM [16], although demonstrating some distortion detection capability, falls short in terms of stable and accurate classification compared to the traditional classification models.

This situation can be attributed to three main reasons:

1. First, remote sensing image quality inspection falls into a specialized domain rather than common knowledge, and the pretraining of large models lacks expertise in this specific field, resulting in insufficient understanding of both images and texts with jargons, such as names of those distortion types.

2. Second, compared to these multimodal model's pretraining dataset consisting of massive image and text data, the dataset used in our study is limited in quantity and diversity.

3. Third, even the largest model we use—VisualGLM [16] has only 6 billion parameters, making it challenging to achieve the same level of answer quality as huge models like GPT-4 [14], which has over 175 billion parameters.

The good news is there is ample room for improvement, including conducting incremental pre-training to instill expertise into the model, using a textual description with additional knowledge, attaching a knowledge base to the model and so on. Appendix D shows a real example of incorporating external knowledge to the question given to VisualGLM [16]. Anyway, due to time and data limitations in this study, a more in-depth investigation is not conducted. Nevertheless, it must be acknowledged that multimodal models possess great potential and are worth exploring in future research.

**5. Conclusion**

This paper introduces a two-step detection method that combines image classification, semantic segmentation, and various image processing techniques, aimed at addressing the inefficiency of traditional methods. It is the first time a two-step deep learning-based method is applied to remote sensing image quality inspection, resulting in a 2.2% mIoU improvement over the one-step detection method, let alone the rule-based methods.

Future work can explore more advanced vision models and incorporate a larger volume of remote sensing image data into the dataset. These enhancements will contribute to improving the generalization and robustness of the model. Additionally, training a multimodal general-purpose model that combines images, texts, and other relevant information can be pursued to further leverage the powerful role of AI in the field of remote sensing.

**Author Contributions**

**Yijiong Yu**: Conceptualization, Methodology, Software, Formal analysis and Writing. **Chang Li** and **Hao Wu**: Resources, Data Curation. **Kang Ran**: Supervision. **Tao Wang**: Project administration and Funding acquisition.

**Data Availability Statement**

The data that support the findings of this study are openly available in Guizhou Remote Sensing Image Quality Inspection Dataset at https://cloud.tsinghua.edu.cn/d/7b3167ee4b8d4242a8d1/, reference number [25].

**Appendix**

**A. Data Collection**

To train a model to detect quality problems in remote sensing images, manually annotated image datasets are necessary. However, while there are many publicly available image quality assessment datasets, they do not meet our task needs, so our own data collection becomes an essential and basic part of our work. This section will introduce how we obtain data to make datasets.

*A.1. Data Source*

The data used in this study is obtained from the Guizhou Surveying and Mapping Product Quality Supervision and Inspection Station and all remote sensing images are taken from Guizhou Province, China, whose climate is cloudy and foggy, and the main terrain is mountainous. The original images are usually in tiff format, with file sizes usually ranging from 300M to 10G and image widths typically greater than 3000 pixels. Each remote sensing image has been manually annotated by professional technicians to identify and mark the regions of the 10 types of distortions. The annotation information for each image is stored in a shapefile (.shp) format, where each distortion region is represented as a polygon, with each vertex of it having latitude and longitude coordinates.

*A.2. Dataset Creation*

Based on the polygon information recorded in the shapefile, and the corresponding remote sensing images in RGB, the detection results are generated using the "draw_polygon" function of scikit-image python package, creating masks that correspond to the inspection results. The mask is a single-channel image where pixels with a value of 0 represent qualified regions (also called background), while other pixel values represent different distortion types. A pixel value of 255 is used to indicate areas that can be ignored, typically representing the black-filled areas outside the range of the remote sensing image. The mapping between pixel values and distortion types is shown in Table 8.

**Table 8.** Mappings between pixel value and category in the dataset.

| Pixel Value | Label |
|---|---|
| 0 | Background |
| 1 | Cloud |
| 2 | Shadow |
| 3 | Stretching Blur Pattern |
| 4 | Blur |
| 5 | Spectral Overflow |
| 6 | Twisted Objects |
| 7 | Stitching Trace |
| 8 | Stitching Misalignment |
| 9 | Stripe Noise |
| 10 | Pixel Missing |
| 255 | (ignored) |

Because the original image size is very huge, to fit the input size of deep learning models, the remote sensing images and their masks are divided into smaller blocks using the same method, whose sizes depend on the resolution of the image and usually set to $512 * 512$. The size of the block should not be too large to ensure that distortion can still be observed with the naked eye after the block is rescale to $512 * 512$. Thus, we create a semantic segmentation dataset containing image and mask pairs.

Furthermore, using the information in the masks, the dataset is further expanded to record which distortion types present in each mask. Additionally, textual descriptions of the images are generated based on certain rules, as outlined in Table 11. The final dataset consists of four columns: image, mask, label set, and textual description, with the data formats being a RGB image, a single-channel mask, a list of label ids, and string respectively.

The dataset contains 2855 realistic remote sensing images with distortion regions (and their masks). To make the distribution of the dataset closer to the sample distribution in real-world environments, 1027 qualified remote sensing images are added as negative samples. Subsequently, the dataset is randomly split into training and testing sets in a 9:1 ratio, while ensuring that the proportions of different image types remain balanced in both sets. The final training set contains 3492 samples, and the testing set contains 390 samples.

*A.3. Data Augmentation*

The occurrence frequencies of quality issues in remote sensing images are highly imbalanced. Figure 8 shows the number of samples containing each distortion type, revealing that certain categories such as Stitching Trace and Spectral Overflow occur much less frequently. As a consequence, the accuracy of classification for these categories will be generally lower.

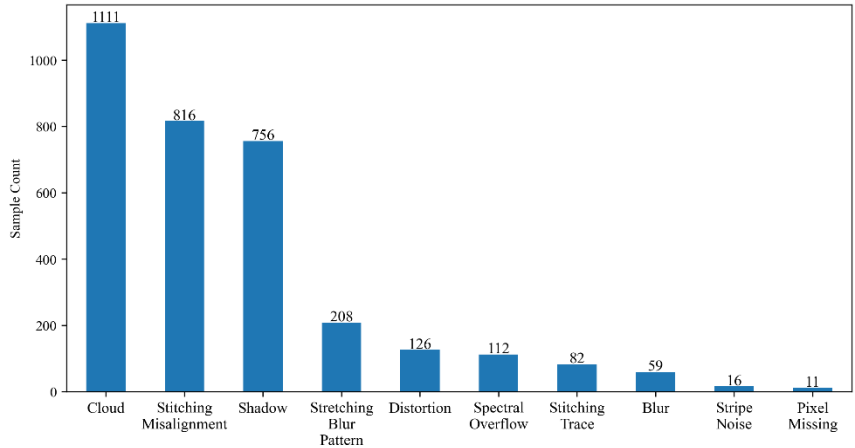

**Figure 8.** Statistics on the number of various types of samples in the dataset.

To address the issue of limited training data, the Stable Diffusion model [26], which has the ability to generate high-quality images based on textual guidance, is introduced in this study to augment data for the Stitching Trace and Spectral Overflow categories. A small set of remote sensing images (approximately 30) along with their corresponding textual descriptions are used to train Stable Diffusion [26] by Lora [24] method. Subsequently, this trained model is employed to generate a large number of similar images, which can be used as new training samples. This approach effectively expands the dataset, improves the performance of image classification, and is more efficient than manually generating data. The textual prompts used during training are in English, as shown in Table 9.

**Table 9.** Correspondence between the text prompt and image content used in Stable Diffusion training.

| Text Prompt | Image Content |
|---|---|
| "Red, yellow, green, or blue patches in satellite images." | Image with Spectral Overflow |
| "Color difference between two parts of Satellite imagery." | Image with Stitching Trace |

Because image generating is not completely controllable, it is necessary to manually filter the available images. We generate a total of 1000 samples first and manually select 200 of them, which could be used for image classification tasks. However, if they are to be used for semantic segmentation tasks, manually annotating masks would still be required, which is laborious. Therefore, the images generated by Stable Diffusion in this study are only applicable for image classification datasets.

Table 10 presents the impact of data augmentation on image classification models. The experimental results indicate that using the Stable Diffusion model for data augmentation further improves the performance of image classification, although it is not that much improvement.

**Table 10.** The effect on classification of data augmentation.

|  | Macro F1 of SwinV2 | Macro F1 of BLIP |
|---|---|---|
| data augmentation | 0.9780 | 0.9720 |
| w/o data augmentation | 0.9719 | 0.9611 |

*A.4. Creation of Image-based Question-Answer Dataset*

The training of an image-text multimodal large model requires image-text pairs or image-based question-answer pairs. Since a single question-answer pair may not fully describe the information in the image, using multiple Q&As to describe an image is needed. While manually crafting question-answer pairs yields higher quality, it is costly. Therefore, in this study, the question-answer pairs are generated based on the information extracted from the masks by using specific mapping rules: The correspondence between the image content and the corresponding template for text Q&A is shown in Table 11, and Table 12, Table 13 and Figure 9 respectively shows how the area size, the seam line type and the position of the distortion region are described by text.

If an answer starts with "no", it is considered a negative sample. Having too many negative samples can adversely affect the model's performance, so finally only 20% of the negative samples are randomly retained. The resulting question-answer dataset comprises 10,457 samples for the training dataset and 1,176 samples for the test dataset. Each sample consists of one image and one Q&A pair, and multiple samples may share the same image.

Please note that our trained model is actually a Chinese question-answering model, and all the text data is in Chinese, as we mainly target Chinese users. However, for the sake of reader comprehension, it has been translated into English.

**Table 11.** The distortions contained in the image and the corresponding template of textual Q&A.

| Model | Task Type | Distortions Present in The Image | Question | Answer |
|---|---|---|---|---|
| BLIP | image classification | any | \ | {Distortion name 1, distortion name 2...} exists in the satellite image. |
|  |  | non-existent | \ | The Satellite image is fully qualified. |
| VisualGLM | image classification | any | Is there a quality problem in this satellite image? | Yes, this satellite image has quality problems, including: {distortion name 1, distortion name 2...}. |
|  |  | non-existent | Is there a quality problem in this satellite image? | No, there is no quality problem in this satellite image. |
|  | detailed description | Region-like distortions | Is there an {distortion name} in this satellite image? | Yes, there is an {distortion name} of {one / many} {large / medium / small / tiny} area in this satellite image, which is mainly located in the {position} of the image |
|  |  | Stitching-like distortions | Is there an {distortion name} in this satellite image? | Yes, there is a trace of {distortion name} in this satellite image, located in the {position} of the image, roughly {horizontal / vertical / oblique / broken line}. |
|  |  | pixel-like distortions | Is there an {distortion name} in this satellite image? | Yes, there is an {one / more} {distortion name} in this satellite image, mainly located in the {position} of the image |

**Table 12.** The proportion of the distortion part to the entire image area (denoted by **a**) corresponds to the text description.

| 0.4＜**a**≤1 | 0.1＜**a**≤0.4 | 0.01＜**a**≤0.1 | 0.0001＜**a**≤0.01 | 0≤**a**≤0.0001 |
|---|---|---|---|---|
| large | medium | small | tiny | not exist |

**Table 13.** The angle $\theta$ between the line and the x-axis corresponds to the text description.

| 0°<$\theta$<30 | 60<$\theta$<90 | 30<$\theta$<60 | Not a straight line |
|---|---|---|---|
| horizontal line | vertical line | oblique line | broken line |

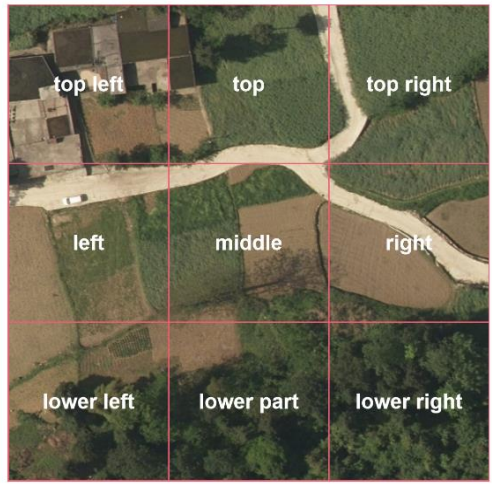

**Figure 9.** The grid where a distortion is located is defined as the grid with the largest area of it. Divide an image into 9 grids, corresponding to 9 different position descriptions, so that the model can roughly locate and describe distortions.

## B. The Pretrained Models Used

The pretrained models used in this article, their download links and parameter amount are shown in Table 14.

**Table 14.** The models used in this paper.

| Task | Model | Download Address | Params |
|---|---|---|---|
| Image classification | Resnet-50 | https://huggingface.co/microsoft/resnet-50 | 25M |
| | SwinV2-base | https://huggingface.co/microsoft/swinv2-base-patch4-window12-192-22k | 104M |
| Image-text multimodality | BLIP-Chinese | https://huggingface.co/IDEA-CCNL/Taiyi-BLIP-750M-Chinese | 750M |
| | VisualGLM | https://huggingface.co/THUDM/VisualGLM-6b | 6B |
| Semantic segmentation | Mobilenet-v2 | https://huggingface.co/google/mobilenet_v2_1.0_224 | 2.6M |
| | Segformer-b5 | https://huggingface.co/nvidia/segformer-b5-finetuned-ade-640-640 | 80M |
| Zero-shot semantic segmentation | CLIPSeg | https://huggingface.co/CIDAS/CLIPSeg-rd64-refined | 144M |
| | GroupVit | https://huggingface.co/nvidia/groupvit-gcc-yfcc | 53M |
| Text to image | stable-diffusion-v1-5 | https://huggingface.co/runwayml/stable-diffusion-v1-5 | 1B |

### C. The Rules of Threshold Segmentation for Region-like Distortions

In the past work, we used to extract cloud and shadow areas by the threshold segmentation method, but other types of distortion were still inspected manually, because they have no feature that can be detected by the threshold method.

The rules for detecting cloud and shadow are shown in Equations (4) and (5), respectively, where R/G/B represents the pixel values of the red/green/blue channel. After extracting the pixels that meet the threshold, a connected component analysis is performed, and components with areas less than 100 pixels are removed.

$$Cloud: \quad R > 150 \; and \; G > 150 \; and \; B > 150 \tag{4}$$

$$Shadow: \quad G > B > R \; and \; 10 < G - R < 35 \; and \; G < 50 \tag{5}$$

### D. An Example of Incorporating External Knowledge into Prompt

When giving VisualGLM [16] a remote sensing image with "spectral overflow" and asking "Is there any spectral overflow in this image?" The model will give the wrong answer of "No". If you add "spectral overflow refers to some red, yellow, blue and green patches in the remote sensing image" to the question, the model will give the correct answer, "Spectral overflow does exist in this image." The origin Chinese dialogue is shown in Figure 10.

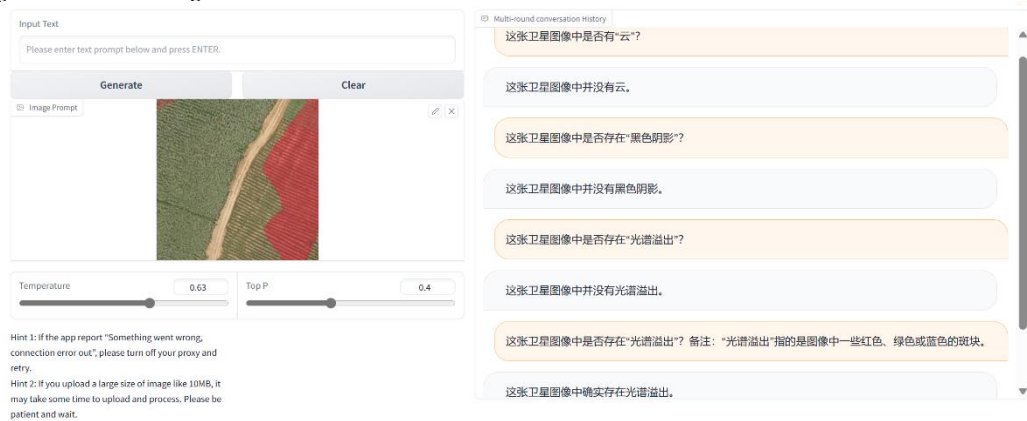

**Figure 10.** VisualGLM corrects its answer by additional information in the question.